\documentclass[11pt]{article}

\usepackage[final]{acl}

\usepackage{times}
\usepackage{latexsym}

\usepackage[T1]{fontenc}

\usepackage[utf8]{inputenc}

\usepackage{microtype}

\usepackage{inconsolata}

\usepackage{graphicx}

\usepackage{algorithm}
\usepackage{algorithmic}
\usepackage{amsmath}

\usepackage{tabularx, booktabs}
\usepackage{hhline}
\usepackage{placeins}
\usepackage{afterpage}
\usepackage{marvosym}
\usepackage{newfloat}
\usepackage{listings}
\DeclareCaptionStyle{ruled}{labelfont=normalfont,labelsep=colon,strut=off} 
\floatstyle{ruled}
\newfloat{listing}{tb}{lst}{}
\floatname{listing}{Listing}
\title{\textsc{SWAN}: \underline{S}emantic \underline{W}atermarking with \underline{A}bstract Mea\underline{n}ing Representation}

\author{
  Ziping Ye\textsuperscript{1}\thanks{\ Equal contribution.}\thanks{\ \Letter\ Corresponding author: \texttt{zipingye@amazon.com}} \quad
  Gourab Dey\textsuperscript{1}\footnotemark[1] \quad
  Christos Christodoulopoulos\textsuperscript{2}\thanks{\ Work done while at Amazon.} \\
  \textbf{Charith Peris\textsuperscript{1}} \quad
  \textbf{Anil Ramakrishna\textsuperscript{3}}\footnotemark[3] \quad
  \textbf{Weitong Ruan\textsuperscript{1}} \\
  \textbf{Aram Galstyan\textsuperscript{1,4}} \quad
  \textbf{Kai-Wei Chang\textsuperscript{1,5}} \quad
  \textbf{Rahul Gupta\textsuperscript{1}} \quad
  \textbf{Ninareh Mehrabi\textsuperscript{3}}\footnotemark[3] \\
  \textsuperscript{1}Amazon \quad
  \textsuperscript{2}Information Commissioner's Office \quad
  \textsuperscript{3}Meta \quad
  \textsuperscript{4}USC \quad
  \textsuperscript{5}UCLA
}

\begin{document}

\maketitle

\begin{abstract}
We introduce SWAN (Semantic Watermarking with Abstract Meaning Representation)\footnote{Code available at \url{https://github.com/amazon-science/SWAN}}, a novel framework that embeds watermark signatures into the semantic structure of a sentence using Abstract Meaning Representation (AMR). In contrast to existing watermarking methods, which typically encode signatures by adjusting token selection preferences during text generation, SWAN embeds the signature directly in the sentence’s semantic representation. As the signature is encoded at the semantic structure level, any paraphrase that preserves meaning automatically preserves the signature. SWAN is training-free: watermark injection is achieved by prompting an LLM to generate sentences guided by a selected AMR template while maintaining contextual coherence, and detection uses an off-the-shelf AMR parser followed by a simple one-proportion z-test. Empirical evaluation on the \textsc{RealNews} benchmark shows SWAN matches state-of-the-art detection performance on unaltered watermarked text, while significantly improving robustness against paraphrasing, increasing detection AUC by up to 13.9 percentage points compared to prior methods. These results demonstrate that SWAN's approach of anchoring watermarks in AMR semantic structures provides a simple, effective, and prompt-based method for robust text provenance verification under paraphrasing, opening new avenues for semantic-level watermarking research.
\end{abstract}

\section{Introduction}
\label{sec:intro}

Recent advances in large language models (LLMs) enable the generation of human-like text that can convincingly mimic genuine human writing, raising significant concerns about misinformation, impersonation, and unauthorized content reuse at scale \citep{xu2024llmsknowrespectcopyright, williams2024largelanguagemodelsconsistently}. To address these risks, \emph{text watermarking} has emerged as a crucial technique, embedding hidden but detectable signatures into generated content, thereby enabling reliable identification of AI-generated text without compromising readability.

Various approaches have been proposed to embed statistical or encrypted patterns into generated text by selecting tokens based on specific underlying rules. However, these token-level methods \citep{kirchenbauer2023watermark, zhao2023provablerobustwatermarkingaigenerated, kirchenbauer2024reliabilitywatermarkslargelanguage, kuditipudi2024robustdistortionfreewatermarkslanguage}  remain highly sensitive to paraphrasing and synonym substitutions, making the watermark easily undetectable after minor text modifications. To address these limitations, recent methods have shifted towards sentence-level watermarking, embedding signals within the semantic embedding space using techniques such as locality-sensitive hashing \citep{hou-etal-2023-semstamp} or k-means clustering \citep{hou-etal-2024-k}. While these sentence-level approaches improve robustness—meaning the watermark remains detectable even after paraphrasing—they can still fail if paraphrasing shifts the sentence embedding.

In this paper, we introduce \textbf{SWAN} (\underline{S}emantic \underline{W}atermarking with \underline{A}bstract Mea\underline{n}ing Representation), a novel watermarking approach that embeds watermark signatures directly into the semantic structure of generated sentences. Unlike previous token-level or embedding-based approaches, SWAN leverages Abstract Meaning Representation (AMR) \citep{banarescu-etal-2013-abstract, regan2024massivemultilingualabstractmeaning}, a graph-based formalism capturing core semantic relationships. By embedding watermark signals at the AMR graph level—for example, requiring a sentence to contain specific combinations of entities and relationships—SWAN ensures that any paraphrase preserving the underlying meaning will inherently retain the watermark. Anchoring the watermark to the semantic structure substantially enhances robustness to paraphrasing attacks, addressing a key limitation of prior methods while maintaining high detectability and text quality.

We validate the effectiveness and robustness of SWAN through extensive empirical evaluation on the \textsc{RealNews} dataset \citep{raffel2023exploringlimitstransferlearning}, a standard benchmark for sentence-level watermarking. Our experiments benchmark SWAN against state-of-the-art token-level and sentence-level watermarking baselines, assessing detectability under various paraphrase attacks. Results demonstrate that SWAN achieves detection accuracy comparable to state-of-the-art sentence-level watermarking methods and outperforms token-level watermarking techniques on unaltered text. More importantly, SWAN significantly enhances robustness under paraphrasing, consistently outperforming existing sentence-level approaches when detecting paraphrased text. Although watermarking generally introduces a trade-off in text quality, SWAN maintains text coherence, fluency, and diversity comparable to existing methods, as evaluated through a thorough LLM-based quality assessment.

To the best of our knowledge, this work represents the first exploration of Abstract Meaning Representation in the text watermarking domain, opening a new technical pathway based on symbolic semantic structures.




\section{Background}
\label{sec:background}

\subsection{Abstract Meaning Representation (AMR)}
\label{subsec:amr}

Abstract Meaning Representation is a graph-based formalism that encodes the core semantic structure of a sentence by abstracting away surface-level syntactic and lexical variations \citep{banarescu-etal-2013-abstract, regan2024massivemultilingualabstractmeaning}. Nodes in the graph represent \emph{concepts} (e.g., entities, events), while edges capture \emph{semantic relations} (e.g., agent, patient, location).

For instance, the sentences below, despite their surface-level differences, share the same core semantic meaning and therefore map to an identical AMR graph \citep{amr-guidelines}:

\begin{itemize}
    \item ``The boy desires the girl to believe him.''
    \item ``The boy desires to be believed by the girl.''
    \item ``The boy has a desire to be believed by the girl.''
    \item ``The boy’s desire is for the girl to believe him.''
\end{itemize}

These sentences correspond to the following AMR graph representation:

\lstdefinestyle{prompt}{basicstyle=\ttfamily,breaklines=true}
\begin{lstlisting}[style=prompt]
(w / want-01
   :ARG0 (b / boy)
   :ARG1 (b2 / believe-01
             :ARG0 (g / girl)
             :ARG1 b))
\end{lstlisting}

\begin{figure}[htbp]
    \centering
    \includegraphics[width=0.65\linewidth]{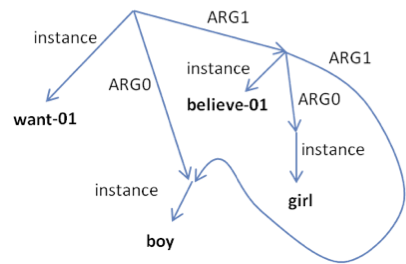}
    \caption{Visualization of an AMR graph capturing semantic equivalences across multiple paraphrases.\protect\footnotemark} 
    \label{fig:amr-example}
\end{figure}

\noindent Figure~\ref{fig:amr-example} provides a visual depiction of this AMR structure, clearly illustrating the semantic relationships captured by the graph representation.

\footnotetext{Figure adapted from \url{https://github.com/amrisi/amr-guidelines/blob/master/graph.png}.}
By representing semantic structures explicitly, AMR provides an effective means for embedding robust watermark signals that remain intact even when sentences undergo significant lexical and syntactic modifications, as long as the core meaning remains unchanged.

\begin{figure*}[htbp]
  \centering
  \includegraphics[width=0.925\textwidth]{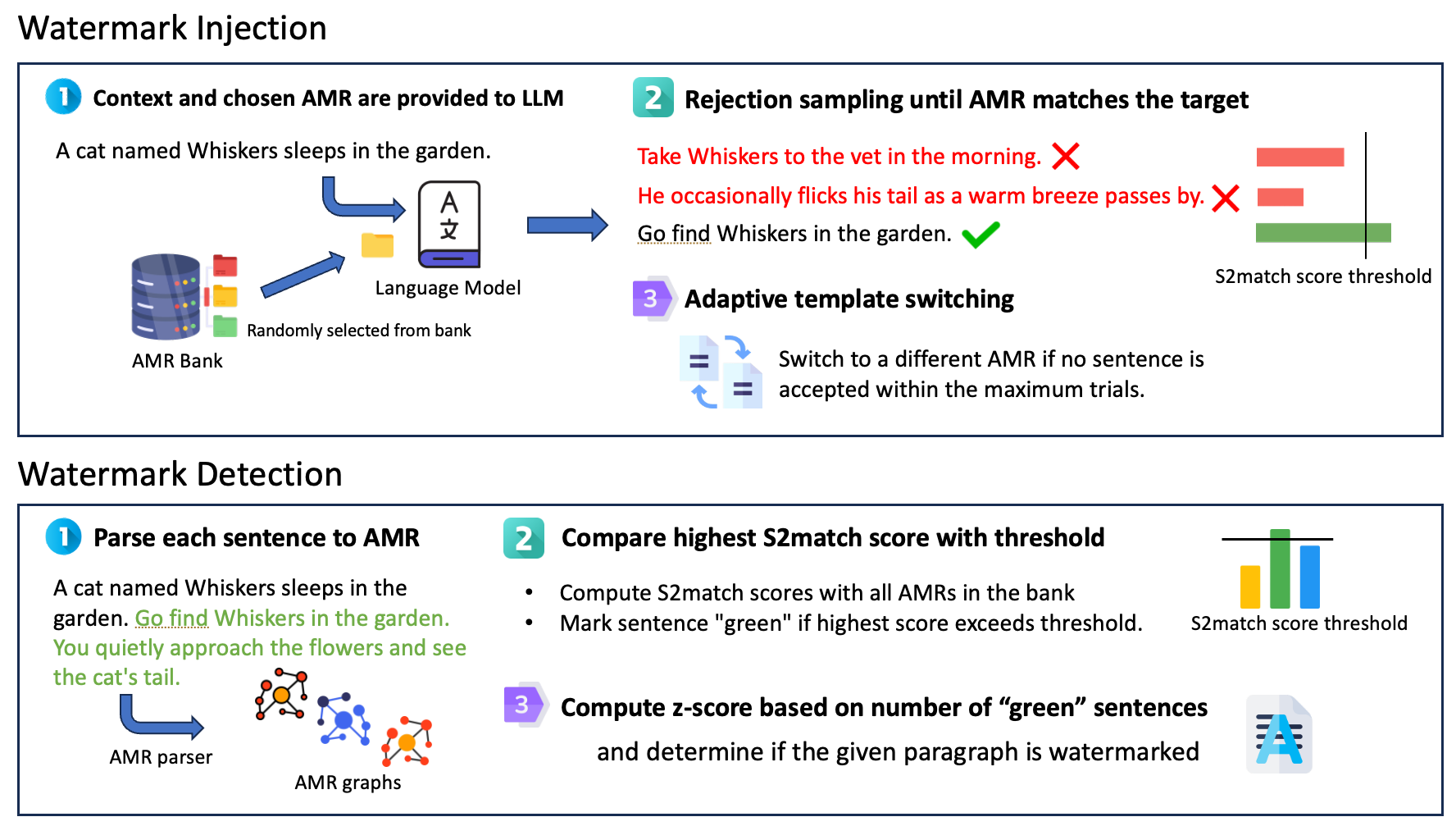}
  \caption{Overview of our proposed framework. In \textbf{watermark injection}, LLM repeatedly samples a sentence until its parsed AMR matches a secret template drawn from the bank; in \textbf{watermark detection}, we parse each sentence of a candidate paragraph, count AMR matches, and apply a $z$-test—because the watermark lives in the AMR graph, any paraphrase that preserves meaning leaves the signal intact.
}
  \label{fig:framework}
\end{figure*}

\subsection{Text Watermarking}
\label{subsec:text_watermarking}
Text watermarking refers to techniques that embed subtle but identifiable signatures into text to allow verification of its provenance or authorship. Depending on the granularity at which the signal is embedded, watermarking methods can be broadly classified into \textit{token-level}, \textit{sentence-level}, and \textit{paragraph-level} approaches.

\paragraph{Token-Level Watermarking.}
Token-level schemes watermark text by subtly biasing the sampling distribution during decoding, so that the hidden statistical signature can later be detected \citep{kirchenbauer2023watermark, kirchenbauer2024reliabilitywatermarkslargelanguage, kuditipudi2024robustdistortionfreewatermarkslanguage, zhao2023provablerobustwatermarkingaigenerated, shi2023redteaminglanguagemodel, dathathri2024synthid}. A widely used design is the \emph{green-list / red-list} framework of \citet{kirchenbauer2023watermark}, which partitions the vocabulary with a secret hash and forces generation to prefer the green tokens, enabling a simple $z$-score detector. Subsequent variants add provable robustness guarantees \citep{zhao2023provablerobustwatermarkingaigenerated}, enforce distortion-free constraints \citep{kuditipudi2024robustdistortionfreewatermarkslanguage}, or optimize the logits with a learned key to balance invisibility and detectability, as in SynthID-Text \citep{dathathri2024synthid}. Although straightforward, token-level methods remain fragile to paraphrasing and other surface edits, and they typically require direct access to model logits—an assumption that may not hold for proprietary LLMs.


\paragraph{Semantic and Sentence-Level Watermarking.}
To enhance robustness against paraphrasing, recent strategies encode signals into the semantic space rather than superficial token statistics. \citet{hou-etal-2023-semstamp} and \citet{hou-etal-2024-k} partition the embedding space using locality-sensitive hashing (LSH) or $k$-means to selectively sample sentences from ``watermarked'' regions. Similarly, \citet{liu2024semanticinvariantrobustwatermark} train a neural network to map context embeddings directly to watermark logits, while \citet{ren2024robustsemanticsbasedwatermarklarge} discretize the continuous embedding space to robustly seed vocabulary partitioning. Although these embedding-based methods better resist small lexical changes, paraphrasing that moves the sentence embedding out of the designated region still degrades detectability.

\paragraph{Paragraph-Level and Post-hoc Watermarking.}
Another line of research explores embedding watermark signals at the paragraph level or injecting them post-generation in a black-box manner. PostMark \citep{chang2024postmark} exemplifies this by first encoding the semantic content of an entire paragraph, then selecting specific ``watermark words'' from a secret embedding table that an instruction-following LLM weaves into the existing text. \emph{Detection} re-computes the expected word list and flags a paragraph when enough of the words (or close semantic neighbors) appear. While practical and model-agnostic, these methods introduce explicit lexical edits, making them potentially detectable or removable by sophisticated adversaries.

Our work diverges from these prior paradigms by embedding explicit watermark signatures directly into the semantic structure of sentences using Abstract Meaning Representation (AMR) graphs. This novel semantic anchoring approach provides robust detection under paraphrasing, addressing key limitations of existing token-level and embedding-based watermarking methods.

\section{Overview of the Framework}
\label{sec:framework}
\textsc{SWAN} embeds a \emph{sentence-level} watermark by guiding each generated sentence to match a template AMR randomly drawn from a secret bank. During generation, sentences are produced conditioned on the accumulated context while adhering to the selected AMR template's semantic structure. Detection parses candidate text back to AMRs, counts template matches, and applies a one-proportion $z$-test over the paragraph. Figure~\ref{fig:framework} summarizes the three components outlined below.

\subsection{AMR Bank Creation}
\label{subsec:amr_bank}

We begin by constructing a \emph{bank} $\mathcal{B}$—a private collection of template AMR graphs that serves as the secret key for both watermark injection and detection. SWAN draws one template from this bank for each sentence to be generated.

\paragraph{Raw corpus.}
We start from MASSIVE-AMR \citep{regan2024massivemultilingualabstractmeaning}\footnote{\url{https://github.com/amazon-science/MASSIVE-AMR}}, which provides $\sim$84\,K AMR graphs for 1,685 information-seeking utterances.

\paragraph{Template AMR.}
We further abstract raw AMRs into \textit{template AMRs} by replacing nouns (e.g., specific named entities) with generic placeholders. This additional level of abstraction removes extraneous lexical details while retaining the underlying semantic structure. For example, a raw AMR node labeled “\texttt{Alice}” might become a placeholder “\texttt{NE},” ensuring that the resulting template AMR focuses on conceptual and relational information rather than domain-specific references. Using these template AMRs confers two main benefits: \textbf{(i) Generality:} A small number of template AMRs can map to a wide variety of sentence instantiations, since names or entities can be filled in later with context-appropriate text. \textbf{(ii) Robustness:} By stripping away specific entity mentions, we decrease the chance that minor lexical edits (e.g., changing “Alice” to “Mary”) will disrupt the watermark’s core semantics.

\paragraph{Selecting AMR Patterns.}
To populate the template bank, we traverse the full MASSIVE-AMR corpus and retain only those graphs that (a) occur within a specified frequency range, defined by a minimum frequency of three and a maximum frequency of twenty, effectively discarding both one-off idiosyncrasies and highly repetitive boilerplate, and (b) contain at least three concept nodes to guarantee non-trivial semantic content.

\subsection{Watermark Injection}
\label{subsec:watermark_injection}

\paragraph{Sentence-Level Guidance.} 
For each sentence to be generated, we draw a template AMR $g$ from the bank and prompt the LLM with two inputs: (1) the accumulated context from previously generated sentences for discourse coherence, and (2) the template $g$ with instructions to generate text adhering to its semantic structure. The accumulated context carries forward the intent from the original user prompt or task specification that initiated the generation. The prompt guides the model to instantiate abstract AMR concepts (e.g., replacing ``NE'' placeholders with appropriate named entities) while maintaining contextual flow and alignment with the user's original intent (full prompt in Appendix \ref{app:gen_prompt}). The candidate sentence is then parsed back into an AMR $\hat g$.

\paragraph{S2match-guided rejection.}
For every candidate sentence generated, we parse its AMR $\hat{g}$ and compute the \textsc{S2match} similarity to the target template $g$—a lightweight metric that averages node- and edge-level $F_{1}$ scores and ranges from 0 (no overlap) to 1 (perfect match) \citep{opitz-etal-2021-amr-sim-metrics}. The sentence is accepted only if $\text{S2match}(\hat{g}, g) \ge \theta_{\text{accept}}$; otherwise we discard it and resample.

\paragraph{Adaptive target switching.}
Because generation is conditioned on running discourse, certain templates may be semantically infeasible in a given context or for a given prompt.  Rather than retrying an incompatible template indefinitely, we impose a small cap on resampling attempts; if the threshold is reached, we simply draw a new template at random from the bank and resume decoding.

\begin{algorithm}[ht!]
\small
\caption{Watermark Injection}
\label{alg:watermark_injection_context}
\begin{algorithmic}[1]
\STATE {\bfseries Input:} AMR bank $\mathcal{B}$, initial sentence $s_0$, number of sentences $N$, max templates $T$, max attempts per template $M$, threshold $\theta_{\text{accept}}$
\STATE $\textit{outputSentences} \gets []$
\STATE $\textit{context} \gets s_0$
\FOR{$i = 1$ \TO $N$}
    \STATE $\textit{acceptedSentence} \gets \textbf{False}$
    \STATE $\textit{chosenText} \gets \text{None}$
    \FOR{$t = 1$ \TO $T$}
        \STATE $\textit{templateAMR} \gets \text{RandomSample}(\mathcal{B})$
        \FOR{$m = 1$ \TO $M$}
            \STATE \textit{generatedText} $\gets$ \text{LLM\_Generate}(\textit{templateAMR}, \textit{context})
            \STATE $\textit{parsedAMR} \gets \text{ParseToAMR}(\textit{generatedText})$
            \STATE $\textit{score} \gets \text{S2match}(\textit{parsedAMR}, \textit{templateAMR})$
            \IF{$\textit{score} \ge \theta_{\mathrm{accept}}$}
                \STATE $\textit{chosenText} \gets \textit{generatedText}$
                \STATE $\textit{acceptedSentence} \gets \textbf{True}$
                \STATE \textbf{break}
            \ENDIF
        \ENDFOR
        \IF{$\textit{acceptedSentence}$} \STATE \textbf{break} \ENDIF
    \ENDFOR
    \IF{not $\textit{acceptedSentence}$}
        \STATE $\textit{chosenText} \gets \textit{generatedText}$ \COMMENT{Fallback to the last attempt}
    \ENDIF
    \STATE $\textit{outputSentences}.\text{append}(\textit{chosenText})$
    \STATE $\textit{context} \gets \textit{context} + \text{" "} + \textit{chosenText}$ \COMMENT{Update context with the accepted sentence}
\ENDFOR
\STATE \textbf{return} $\textit{outputSentences}$
\end{algorithmic}
\end{algorithm}

\subsection{Watermark Detection}
\label{subsec:watermark_detection}

Given a piece of text of unknown provenance, we determine whether it was produced by \textsc{SWAN} in three steps: 
\begin{enumerate}
    \item \textbf{AMR Parsing:} Convert each sentence into an AMR graph using an off-the-shelf AMR parser. 
 \item  \textbf{Pattern Matching:} For the parsed graph \(\hat g\) we compute
          \(\max_{g \in \mathcal{B}}\text{S2match}(\hat g, g)\), the
          best similarity to any template in the bank~\(\mathcal{B}\).
          The sentence is labelled \textit{watermarked} if this score
          exceeds a fixed threshold~\(\theta_{\text{detect}}\). 
 \item \textbf{Paragraph test (one-proportion \(z\)-test).}  
          If \(k\) of the \(n\) sentences in a paragraph are flagged,
          we compute:
          \[
              z \;=\; \frac{\;k - \lambda n\;}
              {\sqrt{\,n\lambda(1-\lambda)\,}},
          \]
          where \(\lambda\) is the expected hit-rate under the null hypothesis:

        \noindent
        $H_{0}:$\\
        \textit{The text is not generated (or written) knowing the secret AMR bank.}
\end{enumerate}

\noindent
Because paraphrasing typically preserves semantic structure, the AMRs remain consistent even when surface forms differ, ensuring that our watermark remains detectable unless meaning has been fundamentally altered.

\begin{algorithm}[ht!]
\small
\caption{Paragraph-Level Watermark Detection}
\label{alg:watermark_detection}
\begin{algorithmic}[1]
\STATE {\bfseries Input:} AMR bank $\mathcal{B}$, detection threshold $\theta_{\text{detect}}$, similarity function $\text{S2match}(\cdot,\cdot)$, $\text{ZscoreTest}(\cdot)$, paragraph $p$
\STATE $\textit{greenCount} \gets 0$
\STATE $\textit{totalSentences} \gets 0$
\FOR{\textbf{each} sentence $s$ in paragraph $p$}
    \STATE $\textit{parsedAMR} \gets \text{ParseToAMR}(s)$
    \STATE $\textit{bestScore} \gets -\infty$
    \FOR{\textbf{each} $\textit{templateAMR}$ in $\mathcal{B}$}
        \STATE $\textit{score} \gets \text{S2match}(\textit{parsedAMR}, \textit{templateAMR})$
        \IF{$\textit{score} > \textit{bestScore}$}
            \STATE $\textit{bestScore} \gets \textit{score}$
        \ENDIF
    \ENDFOR
    \IF{$\textit{bestScore} \ge \theta_{\mathrm{detect}}$}
        \STATE $\textit{greenCount} \gets \textit{greenCount} + 1$
    \ENDIF
    \STATE $\textit{totalSentences} \gets \textit{totalSentences} + 1$
\ENDFOR
\STATE $\textit{greenFrac} \gets \frac{\textit{greenCount}}{\textit{totalSentences}}$
\STATE $\textit{decision} \gets \text{ZscoreTest}(\textit{greenFrac})$
\IF{$\textit{decision} = \textbf{True}$}
    \STATE \textbf{return} "Likely watermarked"
\ELSE
    \STATE \textbf{return} "Not watermarked"
\ENDIF
\end{algorithmic}
\end{algorithm}

\section{Experiments}
\label{sec:experiments}

\subsection{Experimental Setup}
\label{subsec:exp_setup}
\paragraph{Dataset.}
Following the SemStamp baseline, we evaluate on the \textsc{RealNews} subset of the C4 corpus \citep{raffel2023exploringlimitstransferlearning}. \textsc{RealNews} consists of professionally written news articles and has become the de-facto test bed for sentence-level watermarking because its formal style reduces noise from ill-formed user content. We take the first 250 sentences of the subset as our evaluation set.

\paragraph{AMR-bank size.}
A ``bank'' is the private set of template AMRs from which SWAN draws one template per sentence.  Its size $|\mathcal{B}|$ therefore controls the
search space for both injection and detection.  All results reported in Sec.~\ref{sec:experiments} use a  \textbf{50-template bank} ($|\mathcal{B}|{=}50$); Sec.~\ref{sec:bank-size} shows that performance is stable for larger banks up to 800 templates.

\paragraph{Models.}
We use the following models for each component in SWAN:\footnote{For open-source models, we used NVIDIA H100 and A100 GPUs based on availability, and for closed-source models, we used Amazon Bedrock.}
\begin{itemize}
    \item \textbf{Watermark Generation.} All watermark injection is performed with \texttt{DeepSeek-R1-Distill-Qwen-14B} (temperature: 0.6, top\_p: 0.9).\footnote{\url{https://huggingface.co/deepseek-ai/DeepSeek-R1-Distill-Qwen-14B}} SWAN employs rejection sampling capped at 50 trials per sentence (5 attempts per AMR template across up to 10 templates), compared to SemStamp's 100 trials per sentence. 

    \item \textbf{Watermark Detection.} SWAN parses sentences with the
\texttt{parse\_xfm\_bart\_large} pipeline from \textsc{amrlib}\ \citep{amrlib}, a BART-large encoder–decoder trained on AMR-3 (LDC2020T02). 
              
    \item \textbf{Paraphrase Generation.} We create attacks with three paraphrasers—Pegasus, Parrot, and Claude 3.7 Sonnet.

    \item \textbf{Text-Quality Evaluation.} Claude 3.7 Sonnet is used, in a zero-shot setting, to rate coherence, fluency, and diversity of the generated text.

\end{itemize}

\paragraph{Baseline methods.}
We benchmark \textsc{SWAN} against the strongest publicly–available watermarking approaches: \textbf{(i) SynthID-Text.} SynthID-Text \citep{dathathri2024synthid} is a production-ready \emph{token-level} watermark that perturbs the next-token distribution during decoding.
\textbf{(ii) SemStamp.} \emph{SemStamp} partitions the sentence-embedding space with the random-hyperplane through locality-sensitive hashing (LSH) \citep{lsh} and keeps only sentences whose LSH code falls in a secret ``green'' bucket \citep{hou-etal-2023-semstamp}.
\textbf{(iii) k-SemStamp.} k-SemStamp \citep{hou-etal-2024-k} keeps the same accept/reject scheme but swaps the random-hyperplane LSH for $k$-means centroids learned from in-domain text, so regions follow real semantic structure which boosts paraphrase robustness.

\paragraph{Paraphrase attacks.}
Watermark robustness is evaluated under three paraphrasers:
\textbf{(i) Pegasus.} Pegasus \citep{zhang2020pegasuspretrainingextractedgapsentences} is a model fine-tuned for paraphrasing.\footnote{\url{https://huggingface.co/tuner007/pegasus_paraphrase}} \textbf{(ii) Parrot.} Parrot \citep{damodaran2021parrot} is a T5-base paraphraser that seeks high lexical diversity while preserving semantics.
\textbf{(iii) Off-the-shelf LLM.} We use Claude 3.7 in a zero-shot setting instructed to rewrite the given sentence while preserving semantic meaning.

\paragraph{Evaluation metrics.}
We measure two orthogonal aspects:
\emph{(i) watermark quality}—how well the detector distinguishes watermarked text from non-watermarked text—and  
\emph{(ii) text quality} of the watermarked output itself.

\begin{itemize}
    \item \textbf{Watermark detectability.} We report three complementary metrics at the paragraph level (i.e., generate 5 new sentences given one sentence as initial context): (i) \textbf{AUC}, the area under the ROC curve, capturing overall separability between watermarked and non-watermarked text; (ii) \textbf{TPR@1\%} and (iii) \textbf{TPR@5\%}, the true positive rates at fixed false positive rates of 1\% and 5\%, which reflect detector performance in the low-FPR regime most relevant to deployment.
    \item \textbf{Text quality.}  We use Claude 3.7 as the judge and record scores on three axes: \emph{Coherence}, \emph{Fluency}, and \emph{Diversity}. Detailed definitions of these quality dimensions and the complete evaluation prompt are provided in Appendix~\ref{app:quality_prompt}.
\end{itemize}

\subsection{Detection Results}
\label{subsec:detection_results}

\paragraph{Baseline performance (no paraphrase).}
Table~\ref{tab:main} reports paragraph-level AUC on the \textsc{RealNews} set without any rewriting. \textsc{SWAN} matches the sentence-level competitor and outperforms the token-level SynthID watermark, demonstrating that our semantic approach entails no loss of raw detectability.

\begin{table}[t]
\centering
\begin{tabular}{l@{\hspace{1em}}c@{\hspace{0.8em}}c@{\hspace{0.8em}}c}
\toprule
\textbf{Method} & \textbf{AUC} & \textbf{TPR@1\%} & \textbf{TPR@5\%} \\
\midrule
SynthID              & 97.0 & 64.8 & 84.8 \\
SemStamp             & \textbf{99.4} & \textbf{96.8} & \textbf{100}  \\
\textit{k}-SemStamp  & 99.1 & \textbf{96.8} & 96.4 \\
\textsc{SWAN}        & 99.1 & 91.6 & 97.6 \\
\bottomrule
\end{tabular}
\caption{Detection accuracy (\%) on \textsc{RealNews} sentences without paraphrasing. AUC is the area under the ROC curve; TPR@1\% and TPR@5\% are the true positive rates at 1\% and 5\% false positive rates, respectively. \textsc{SWAN} performs comparably to the sentence-level baseline and outperforms the token-level SynthID, showing that its semantic anchoring retains full detectability in the easiest setting.}
\label{tab:main}
\end{table}

\paragraph{Robustness to paraphrasing.}
Table~\ref{tab:para} shows the same evaluation after applying three paraphrasers. SynthID is omitted as SWAN outperforms it even in the non-paraphrasing scenario and its token-level cues are wiped out by even mild rewriting. Across all attacks, \textsc{SWAN} attains the highest AUC, confirming that anchoring the watermark in AMR graphs yields markedly stronger resilience than embedding-partition methods.

\begin{table*}[t]
\centering
\small
\setlength{\tabcolsep}{4pt}
\begin{tabular}{l@{\hspace{0.8em}}c@{\hspace{0.8em}}c@{\hspace{0.8em}}c}
\toprule
\textbf{Method}
  & \textbf{Pegasus}
  & \textbf{Parrot}
  & \textbf{Claude} \\
  & \small{AUC\,/\,TPR@1\%\,/\,TPR@5\%}
  & \small{AUC\,/\,TPR@1\%\,/\,TPR@5\%}
  & \small{AUC\,/\,TPR@1\%\,/\,TPR@5\%} \\
\midrule
SemStamp
  & 97.6\,/\,87.2\,/\,\textbf{97.6}
  & 94.8\,/\,69.2\,/\,97.6
  & 84.4\,/\,36.8\,/\,84.8 \\
\textit{k}-SemStamp
  & 97.3\,/\,\textbf{88.8}\,/\,88.4
  & 92.8\,/\,68.0\,/\,66.8
  & 87.6\,/\,53.6\,/\,53.2 \\
\textsc{SWAN}
  & \textbf{98.1}\,/\,81.2\,/\,92.8
  & \textbf{97.5}\,/\,\textbf{82.0}\,/\,\textbf{92.4}
  & \textbf{98.3}\,/\,\textbf{86.0}\,/\,\textbf{95.2} \\
\bottomrule
\end{tabular}
\caption{Detection performance (\%) under paraphrase attacks, reported as AUC\,/\,TPR@1\%\,/\,TPR@5\%. Across all three paraphrasers, \textsc{SWAN} yields the strongest results, with the largest margins under the zero-shot LLM paraphraser (Claude).}
\label{tab:para}
\end{table*}

\subsection{Effect of AMR-Bank Size}
\label{sec:bank-size}

\begin{table}[!h]
  \centering
  \setlength\tabcolsep{8pt}
  \begin{tabular}{cc}
    \toprule
    \textbf{Bank Size ($|{\mathcal B}|$)} & \textbf{AUC} \\
    \midrule
      50   & 99.1 \\
      100  & 98.7 \\
      500  & 98.4 \\
      800  & \textbf{99.3} \\
    \bottomrule
  \end{tabular}
  \caption{Ablation on the AMR-bank size used during detection, reported as AUC (\%) on unaltered (non-paraphrased) watermarked text. While performance remains high throughout, a bank of 800 AMRs yields the strongest AUC.}
  \label{tab:bank-size}
\end{table}

\begin{table*}[t]
\centering
\small
\renewcommand{\arraystretch}{1.1}
\setlength{\tabcolsep}{5pt}
\begin{tabular}{lcccc}
\toprule
\textbf{Metric} &
\textbf{No Watermark} &
\textbf{SWAN} &
\textbf{SemStamp} &
\textbf{k-SemStamp} \\
\midrule
\textbf{Coherence} &
  $4.19 \pm 1.19\, (0\text{--}5)$ &
  $3.72 \pm 1.17\, (1\text{--}5)$ &
  $3.69 \pm 1.17\, (0\text{--}5)$ &
  $3.70 \pm 1.16\, (1\text{--}5)$ \\[3pt]
\textbf{Fluency} &
  $4.14 \pm 1.21\, (0\text{--}5)$ &
  $3.87 \pm 1.19\, (1\text{--}5)$ &
  $3.85 \pm 1.20\, (0\text{--}5)$ &
  $3.85 \pm 1.20\, (1\text{--}5)$ \\[3pt]
\textbf{Diversity} &
  $3.99 \pm 0.97\, (0\text{--}4.8)$ &
  $3.01 \pm 0.96\, (1\text{--}5)$ &
  $2.99 \pm 0.96\, (0\text{--}4.7)$ &
  $3.00 \pm 0.95\, (1\text{--}4.8)$ \\
\bottomrule
\end{tabular}
\caption{LLM-based text-quality scores (\(0\text{--}5\)) for un-watermarked text and three watermarking schemes, reported as \(\text{Mean} \pm \text{Std}\) \((\text{Min--Max})\).  Relative to the ``No Watermark’’ baseline, all watermarking incurs a modest quality drop, but \textsc{SWAN} matches the sentence-level baselines (SemStamp and \(k\)-SemStamp) while offering stronger paraphrase robustness.}
\label{tab:text_quality_eval}
\end{table*}

\begin{table*}[h]
\small
\centering
\renewcommand{\arraystretch}{1.15}

\begin{tabularx}{\linewidth}{|>{\raggedright\arraybackslash}p{2cm}|X|}
\hhline{|=|=|}
\multicolumn{2}{|c|}{\textbf{SWAN}} \\
\hhline{|=|=|}
\textbf{High-quality} &
\textit{Whitehaven Coal is expected to consider a proposal. Whitehaven Coal’s proposal is expected to be voted on by activist shareholders in record numbers. Vote for Whitehaven Coal. Vote for the proposal to demand that Whitehaven Coal align its strategy with the Paris climate agreement.} \\ \hhline{|=|=|}
\textbf{Low-quality} &
\textit{How does John feel about it? You feel happy. John feels it. John feels happiness about it. John feels happiness.} \\ 
\hhline{|=|=|}
\multicolumn{2}{|c|}{\textbf{SemStamp}} \\
\hhline{|=|=|}
\textbf{High-quality} &
\textit{California is lifting its drought emergency for most of the state after a winter of record rain and snowfall that followed a five-year dry spell. Gov. Jerry Brown’s office announced Friday that his state is now in a drought disaster, with nearly half the state under water restrictions.} \\ \hhline{|=|=|}
\textbf{Low-quality}  &
\textit{WASHINGTON (Reuters) - U.S. Treasury Secretary Jack Lew on Sunday warned Congress against manufacturing a crisis over federal spending in the months ahead, as looming deadlines set the stage for a repeat. of the debt ceiling fights that roiled the country last year. The devil is in the details. Lew told CNN’s State of the Union.} \\ 
\hhline{|=|=|}
\multicolumn{2}{|c|}{\textbf{k-SemStamp}} \\
\hhline{|=|=|}
\textbf{High-quality} &
\textit{A Charlotte cancer patient and his Oncologist just finished one of the most iconic marathons in the country. Wayne Moorehead continues to inspire as he learns to walk without help at TIRR Memorial?, in his hometown of Charlotte, North Carolina. Watch a special interview with him below. This video was originally published on October 20, 2013.} \\ \hhline{|=|=|}
\textbf{Low-quality}  &
\textit{Sports Academy has found “The Mamba Mentality.” The 96,000-square-foot training facility in Thousand Oaks announced a partnership with Kobe Bryant. The 96,000-square-foot holistic athletic training facility in Thousand Oaks announced a partnership with?Kobe!Bryant!That is a first for the facility?s history.} \\ 
\hhline{|=|=|}
\end{tabularx}

\caption{Illustrative paragraphs for each watermarking method.  
Using the LLM-judged coherence, fluency, and diversity metrics (§\ref{subsec:text_quality_evaluation}), we select one example that scores toward the higher end (\emph{High-quality}) and another toward the lower end (\emph{Low-quality}) to visualise the range of output quality.}
\label{tab:qualitative}
\end{table*}

We further ablate the \emph{bank size}, i.e., the number of AMRs in $\mathcal{B}$. Intuitively, a larger bank provides greater semantic coverage, so the watermark can be embedded across more diverse AMR patterns. However, this also raises the possibility of false positives, since any human-written sentence that incidentally aligns with one of many templates may be flagged. Table~\ref{tab:bank-size} shows that while a bank of 800 AMRs yields the highest AUC (99.3\%), even a much smaller bank (50 AMRs) maintains strong performance (99.1\%). This indicates that \textsc{SWAN} is robust to different bank sizes, allowing practitioners to balance coverage and efficiency when deploying the watermark.

\subsection{Text Quality Evaluation}
\label{subsec:text_quality_evaluation}

To assess whether embedding a watermark affects generated text quality, we used a large language model (LLM) as an automated “reference-free” judge \citep{chen2023exploringuselargelanguage}, rating paragraphs on a $[0,5]$ scale along three dimensions: \textbf{(i) Coherence} (logical organization and clarity), \textbf{(ii) Fluency} (grammatical correctness and readability), and \textbf{(iii) Diversity} (variety of vocabulary and sentence structures).

Table~\ref{tab:text_quality_eval} compares text generated without watermarking (``No Watermark'') against three watermarking methods: \textbf{SWAN}, \textbf{SemStamp}, and \textbf{k-SemStamp}. Results indicate that while watermarking introduces a slight decrease in quality scores across all dimensions compared to the un-watermarked baseline, \textsc{SWAN} performs comparably to existing sentence-level watermarking methods. Thus, SWAN achieves strong paraphrase robustness (as demonstrated in \S\ref{subsec:detection_results}) without introducing additional penalties in text coherence, fluency, or diversity relative to current state-of-the-art approaches.

\subsection{Qualitative Examples}
\label{sec:qualitative}

Table~\ref{tab:qualitative} provides illustrative examples of paragraphs generated by each watermarking method (SWAN, SemStamp, and k-SemStamp). For each method, we present one \emph{High-quality} example and one \emph{Low-quality} example based on the LLM-judged coherence, fluency, and diversity scores (\S\ref{subsec:text_quality_evaluation}), highlighting the typical range of output quality observed in practice.

\subsection{Sampling Efficiency}
\label{subsec:sampling_efficiency}

\begin{figure}[!htbp]
\centering
\includegraphics[width=\columnwidth]{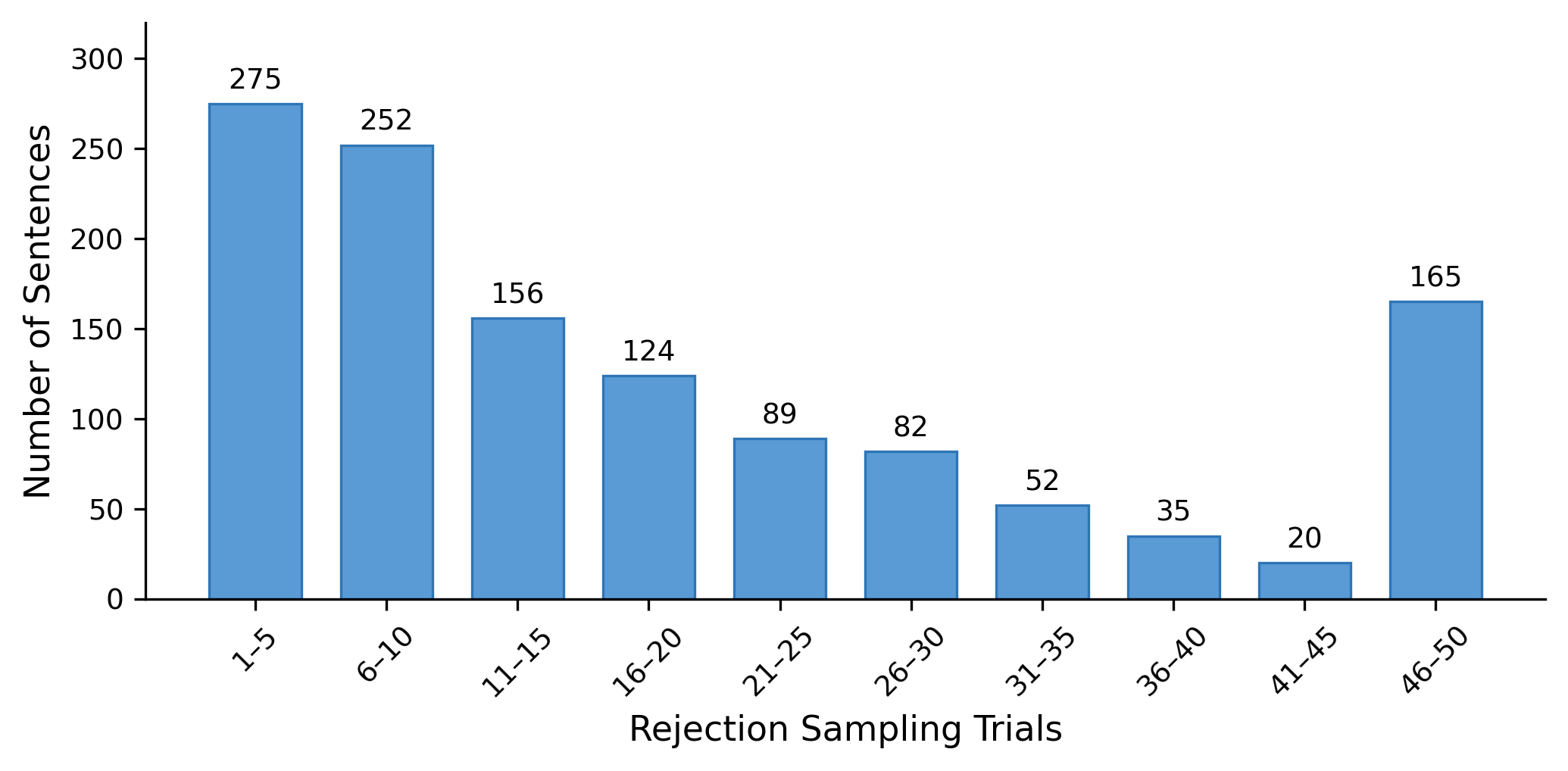}
\caption{Distribution of rejection sampling trials per sentence across 1{,}250 generated sentences (250 samples $\times$ 5 sentences).}
\label{fig:sampling_efficiency}
\end{figure}

Like other sentence‑level algorithms, SWAN relies on rejection sampling. Across 1{,}250 generated sentences (250\,$\times$\,5) we measured a mean of 17.7 trials before acceptance, versus 13.8 trials for SemStamp. Figure~\ref{fig:sampling_efficiency} shows the full trial distribution: 42\% of sentences are accepted within 10 trials and 54\% within 15, confirming that the majority of generations converge quickly. The spike at 46--50 trials corresponds to sentences that exhausted the budget, suggesting that certain AMR templates are harder to satisfy in context; future work on context-aware template selection could reduce these cases and further improve efficiency. While the overall overhead is modest, it is offset by SWAN's significantly enhanced robustness to paraphrasing—our primary contribution—achieving up to 13.9 percentage points higher detection AUC (Table~\ref{tab:para}). SWAN remains \emph{training-free} and works via pure prompting and AMR parsing.


\section{Related Work}
\label{sec:related_work}

\paragraph{Other semantic signatures.}
Early linguistic steganography embeds bits by \emph{rewiring surface structure}. \citet{atallah2001natural} propose two algorithms that hide information in the \textit{parse tree} of selected sentences, modifying syntactic branches and then regenerating fluent text with an NLG system.  More recently, \citet{yoo-etal-2023-robust} advance to a \emph{multi-bit} framework that encodes a larger payload in \textit{invariant features}—lexical or syntactic patterns shown empirically to survive minor text corruption.  Complementary to these text-side schemes, \citet{gu2022watermarking} ``plant'' a watermark \emph{inside} the model itself by backdooring PLM weights with rare-word or composite triggers; ownership is verified by querying the model rather than analyzing the generated text.

\paragraph{Authorship attribution.}
Stylometric methods seek to infer the writer’s identity from persistent linguistic habits rather than from an injected signal.
Classic surveys document feature-rich classifiers that use function-word counts, character $n$-grams, and syntactic cues
\citep{stamatatos2009survey,koppel2009computational}. Neural models learn style embeddings end-to-end, from continuous character $n$-grams \citep{sari-etal-2017-continuous} to the contrastively-trained \textit{PART} Transformer \citep{huertas-tato-etal-2022-part}. Because these approaches rely on labelled samples for every author and can be defeated by paraphrase or adversarial style obfuscation \citep{afroz2012adversarial}, they do not provide a verifiable yes/no provenance test.

\section{Conclusion}
\label{sec:conclusion}

We introduced \textbf{SWAN}, a novel semantic watermarking framework that embeds robust, paraphrase-resistant signals into AMR representations. By guiding generated text to align with curated AMR templates, \textsc{SWAN} remains detectable under paraphrasing as long as core semantic relationships are preserved. Empirically, it outperforms token-level and embedding-based watermark baselines in paraphrase robustness, while maintaining strong text fluency and naturalness.

\paragraph{Future Directions.}
We believe this work opens new directions for semantic-level watermarking research. Improving sampling efficiency represents a key practical direction, potentially through heuristic template selection, AMR-aware language model fine-tuning, or learned template-context matching to reduce rejection sampling overhead. Beyond efficiency, promising research avenues include exploring domain-specific AMR adaptations, improving parser robustness, and strengthening detection against advanced adversarial rewrites. Additionally, investigating how watermarking methods in general—including our AMR-based approach—might impact the reasoning capabilities of LLMs across different task domains represents an important area for future exploration. While \textsc{SWAN} currently focuses on the sentence level, multi-sentence attacks—such as splitting or merging sentences to alter AMR structure—represent a natural next challenge that could be addressed by extending to paragraph-level or multi-sentence AMRs. Another intriguing direction lies in using \emph{AMR subgraphs} as watermark signals, offering additional resilience when adversaries deliberately reorganize semantic structures.

\section*{Limitations}
\label{sec:limitations}

Our detection relies on the quality of the AMR parser, which can occasionally misparse text or introduce random variations. Such errors may reduce recall by pushing a genuine \textsc{SWAN} sentence below the similarity threshold, or increase false positives if they inflate similarity to a template. 


We focused on English news text for evaluation, where well-trained AMR parsers and public corpora are available. In domains with highly specialized jargon or in languages with limited AMR resources, accuracy and detectability might degrade, necessitating more domain-specific or multilingual AMR tools.

In our framework, the AMR bank $\mathcal{B}$ functions as the private key; its secrecy is required to prevent adversary detection. Future work may explore enhancing this security by integrating cryptographic protocols to manage template selection, similar to secret-key hashing in token-level schemes.

\section*{Ethical Considerations}
\label{sec:ethical_consideration}
Watermarking systems, including ours, come with both benefits and risks. On the one hand, they can help combat large-scale misinformation by allowing credible attribution of AI-generated content. On the other hand, false positives could arise if a text happens to align closely with the AMR templates by chance or if the AMR parser errs, potentially mislabeling human-authored text as AI-generated. This risk is especially salient in low-resource languages or domains where AMR parsers are less accurate, raising concerns about fairness and potential bias. Furthermore, mandated watermarking in certain contexts could impede free expression or privacy if applied too broadly. Finally, while our method robustly identifies paraphrases, adversaries may develop sophisticated rewriting strategies to remove the watermark, highlighting the possibility of an ongoing arms race. We encourage responsible use, transparency about these limitations, and continued research to refine watermarking techniques and their governance.

\section*{Acknowledgments}
We thank Dipika Khullar for her contributions to the text quality evaluation experiments. We also thank Richard Zemel for valuable discussions during the course of this work.

\bibliography{custom}

\begin{thebibliography}{31}
\providecommand{\natexlab}[1]{#1}

\bibitem[{Atallah et~al.(2001)Atallah, Raskin, Crogan, Hempelmann, Kerschbaum, Mohamed, and Naik}]{atallah2001natural}
Mikhail~J. Atallah, Victor Raskin, Michael Crogan, Christian Hempelmann, Florian Kerschbaum, Dina Mohamed, and Sanket Naik. 2001.
\newblock Natural language watermarking: Design, analysis, and a proof-of-concept implementation.
\newblock In \emph{Proceedings of the 4th International Workshop on Information Hiding}, IHW '01, page 185–199, Berlin, Heidelberg. Springer-Verlag.

\bibitem[{Banarescu et~al.(2013)Banarescu, Bonial, Cai, Georgescu, Griffitt, Hermjakob, Knight, Koehn, Palmer, and Schneider}]{banarescu-etal-2013-abstract}
Laura Banarescu, Claire Bonial, Shu Cai, Madalina Georgescu, Kira Griffitt, Ulf Hermjakob, Kevin Knight, Philipp Koehn, Martha Palmer, and Nathan Schneider. 2013.
\newblock \href {https://aclanthology.org/W13-2322/} {{A}bstract {M}eaning {R}epresentation for sembanking}.
\newblock In \emph{Proceedings of the 7th Linguistic Annotation Workshop and Interoperability with Discourse}, pages 178--186, Sofia, Bulgaria. Association for Computational Linguistics.

\bibitem[{Banarescu et~al.(2019)Banarescu, Bonial, Cai, Georgescu, Griffitt, Hermjakob, Knight, Koehn, Palmer, and Schneider}]{amr-guidelines}
Laura Banarescu, Claire Bonial, Shu Cai, Madalina Georgescu, Kira Griffitt, Ulf Hermjakob, Kevin Knight, Philipp Koehn, Martha Palmer, and Nathan Schneider. 2019.
\newblock Abstract meaning representation ({AMR}) 1.2.6 specification.
\newblock \url{https://github.com/amrisi/amr-guidelines/blob/master/amr.md}.
\newblock Accessed: 2025-05-12.

\bibitem[{Brennan et~al.(2012)Brennan, Afroz, and Greenstadt}]{afroz2012adversarial}
Michael Brennan, Sadia Afroz, and Rachel Greenstadt. 2012.
\newblock \href {https://doi.org/10.1145/2382448.2382450} {Adversarial stylometry: Circumventing authorship recognition to preserve privacy and anonymity}.
\newblock \emph{ACM Trans. Inf. Syst. Secur.}, 15(3).

\bibitem[{Chang et~al.(2024)Chang, Krishna, Houmansadr, Wieting, and Iyyer}]{chang2024postmark}
Yapei Chang, Kalpesh Krishna, Amir Houmansadr, John Wieting, and Mohit Iyyer. 2024.
\newblock \href {https://arxiv.org/abs/2406.14517} {Postmark: A robust blackbox watermark for large language models}.
\newblock \emph{Preprint}, arXiv:2406.14517.

\bibitem[{Chen et~al.(2023)Chen, Wang, Jiang, Shi, and Xu}]{chen2023exploringuselargelanguage}
Yi~Chen, Rui Wang, Haiyun Jiang, Shuming Shi, and Ruifeng Xu. 2023.
\newblock \href {https://arxiv.org/abs/2304.00723} {Exploring the use of large language models for reference-free text quality evaluation: An empirical study}.
\newblock \emph{Preprint}, arXiv:2304.00723.

\bibitem[{Damodaran(2021)}]{damodaran2021parrot}
Prithiviraj Damodaran. 2021.
\newblock Parrot paraphraser.
\newblock \url{https://github.com/PrithivirajDamodaran/Parrot_Paraphraser}.

\bibitem[{Dathathri et~al.(2024)Dathathri, See, Ghaisas, H{\"{u}}rlimann, Walker, Bartoldson, Mukherjee, Sen, Bansal, Bhasin, Munn, Korotkevich, Singh, Mensink, Hennessey, Venkateswaran, Bichsel, Cooijmans, Ghahramani, Sopyla, Shklovski, Burgess, Gowal, Hassabis, and Kohli}]{dathathri2024synthid}
Sumanth Dathathri, Abigail See, Shubham Ghaisas, Pierre{-}Sacha H{\"{u}}rlimann, Jacob Walker, Brian Bartoldson, Rohan Mukherjee, Aditya Sen, Varun Bansal, Rohan Bhasin, Michael~A. Munn, Alexey Korotkevich, Rishabh Singh, Thomas Mensink, James Hennessey, Nisanth Venkateswaran, Benjamin Bichsel, Thomas Cooijmans, Zoubin Ghahramani, and 6 others. 2024.
\newblock \href {https://doi.org/10.1038/s41586-024-08025-4} {Scalable watermarking for identifying large language model outputs}.
\newblock \emph{Nature}, 634(8035):818--823.

\bibitem[{Gu et~al.(2023)Gu, Huang, Zheng, Chang, and Hsieh}]{gu2022watermarking}
Chenxi Gu, Chengsong Huang, Xiaoqing Zheng, Kai-Wei Chang, and Cho-Jui Hsieh. 2023.
\newblock \href {https://arxiv.org/abs/2210.07543} {Watermarking pre-trained language models with backdooring}.
\newblock \emph{Preprint}, arXiv:2210.07543.

\bibitem[{Hou et~al.(2023)Hou, Zhang, He, Chuang, Wang, Shen, Van~Durme, Khashabi, and Tsvetkov}]{hou-etal-2023-semstamp}
Abe~Bohan Hou, Jingyu Zhang, Tianxing He, Yung-Sung Chuang, Hongwei Wang, Lingfeng Shen, Benjamin Van~Durme, Daniel Khashabi, and Yulia Tsvetkov. 2023.
\newblock \href {https://arxiv.org/abs/2310.03991} {{SemStamp}: A semantic watermark with paraphrastic robustness for text generation}.
\newblock In \emph{Annual Conference of the North American Chapter of the Association for Computational Linguistics}.

\bibitem[{Hou et~al.(2024)Hou, Zhang, Wang, Khashabi, and He}]{hou-etal-2024-k}
Abe~Bohan Hou, Jingyu Zhang, Yichen Wang, Daniel Khashabi, and Tianxing He. 2024.
\newblock \href {https://doi.org/10.18653/v1/2024.findings-acl.98} {{\textit{k}-SemStamp}: A clustering-based semantic watermark for detection of machine-generated text}.
\newblock In \emph{Findings of the Association for Computational Linguistics: ACL 2024}, pages 1706--1715, Bangkok, Thailand. Association for Computational Linguistics.

\bibitem[{Huertas-Tato et~al.(2022)Huertas-Tato, Martin, and Camacho}]{huertas-tato-etal-2022-part}
Javier Huertas-Tato, Alejandro Martin, and David Camacho. 2022.
\newblock \href {https://arxiv.org/abs/2209.15373} {Part: Pre-trained authorship representation transformer}.
\newblock \emph{Preprint}, arXiv:2209.15373.

\bibitem[{Indyk and Motwani(1998)}]{lsh}
Piotr Indyk and Rajeev Motwani. 1998.
\newblock \href {https://doi.org/10.1145/276698.276876} {Approximate nearest neighbors: towards removing the curse of dimensionality}.
\newblock In \emph{Proceedings of the Thirtieth Annual ACM Symposium on Theory of Computing}, STOC '98, page 604–613, New York, NY, USA. Association for Computing Machinery.

\bibitem[{Jascob(2020)}]{amrlib}
Brad Jascob. 2020.
\newblock \href {https://github.com/bjascob/amrlib} {amrlib: A python library that makes amr parsing, generation and visualization simple}.
\newblock Accessed: 2025-05-12.

\bibitem[{Kirchenbauer et~al.(2024{\natexlab{a}})Kirchenbauer, Geiping, Wen, Katz, Miers, and Goldstein}]{kirchenbauer2023watermark}
John Kirchenbauer, Jonas Geiping, Yuxin Wen, Jonathan Katz, Ian Miers, and Tom Goldstein. 2024{\natexlab{a}}.
\newblock \href {https://arxiv.org/abs/2301.10226} {A watermark for large language models}.
\newblock \emph{Preprint}, arXiv:2301.10226.

\bibitem[{Kirchenbauer et~al.(2024{\natexlab{b}})Kirchenbauer, Geiping, Wen, Shu, Saifullah, Kong, Fernando, Saha, Goldblum, and Goldstein}]{kirchenbauer2024reliabilitywatermarkslargelanguage}
John Kirchenbauer, Jonas Geiping, Yuxin Wen, Manli Shu, Khalid Saifullah, Kezhi Kong, Kasun Fernando, Aniruddha Saha, Micah Goldblum, and Tom Goldstein. 2024{\natexlab{b}}.
\newblock \href {https://arxiv.org/abs/2306.04634} {On the reliability of watermarks for large language models}.
\newblock \emph{Preprint}, arXiv:2306.04634.

\bibitem[{Koppel et~al.(2009)Koppel, Schler, and Argamon}]{koppel2009computational}
Moshe Koppel, Jonathan Schler, and Shlomo Argamon. 2009.
\newblock \href {https://doi.org/10.1002/asi.20961} {Computational methods in authorship attribution}.
\newblock \emph{Journal of the American Society for Information Science and Technology}, 60(1):9--26.

\bibitem[{Kuditipudi et~al.(2024)Kuditipudi, Thickstun, Hashimoto, and Liang}]{kuditipudi2024robustdistortionfreewatermarkslanguage}
Rohith Kuditipudi, John Thickstun, Tatsunori Hashimoto, and Percy Liang. 2024.
\newblock \href {https://arxiv.org/abs/2307.15593} {Robust distortion-free watermarks for language models}.
\newblock \emph{Preprint}, arXiv:2307.15593.

\bibitem[{Liu et~al.(2024)Liu, Pan, Hu, Meng, and Wen}]{liu2024semanticinvariantrobustwatermark}
Aiwei Liu, Leyi Pan, Xuming Hu, Shiao Meng, and Lijie Wen. 2024.
\newblock \href {https://arxiv.org/abs/2310.06356} {A semantic invariant robust watermark for large language models}.
\newblock \emph{Preprint}, arXiv:2310.06356.

\bibitem[{Opitz et~al.(2020)Opitz, Parcalabescu, and Frank}]{opitz-etal-2021-amr-sim-metrics}
Juri Opitz, Letitia Parcalabescu, and Anette Frank. 2020.
\newblock \href {https://doi.org/10.1162/tacl_a_00329} {{AMR} similarity metrics from principles}.
\newblock \emph{Transactions of the Association for Computational Linguistics}, 8:522--538.

\bibitem[{Raffel et~al.(2023)Raffel, Shazeer, Roberts, Lee, Narang, Matena, Zhou, Li, and Liu}]{raffel2023exploringlimitstransferlearning}
Colin Raffel, Noam Shazeer, Adam Roberts, Katherine Lee, Sharan Narang, Michael Matena, Yanqi Zhou, Wei Li, and Peter~J. Liu. 2023.
\newblock \href {https://arxiv.org/abs/1910.10683} {Exploring the limits of transfer learning with a unified text-to-text transformer}.
\newblock \emph{Preprint}, arXiv:1910.10683.

\bibitem[{Regan et~al.(2024)Regan, Wein, Baker, and Monti}]{regan2024massivemultilingualabstractmeaning}
Michael Regan, Shira Wein, George Baker, and Emilio Monti. 2024.
\newblock \href {https://arxiv.org/abs/2405.19285} {Massive multilingual abstract meaning representation: A dataset and baselines for hallucination detection}.
\newblock \emph{Preprint}, arXiv:2405.19285.

\bibitem[{Ren et~al.(2024)Ren, Xu, Liu, Cui, Wang, Yin, and Tang}]{ren2024robustsemanticsbasedwatermarklarge}
Jie Ren, Han Xu, Yiding Liu, Yingqian Cui, Shuaiqiang Wang, Dawei Yin, and Jiliang Tang. 2024.
\newblock \href {https://arxiv.org/abs/2311.08721} {A robust semantics-based watermark for large language model against paraphrasing}.
\newblock \emph{Preprint}, arXiv:2311.08721.

\bibitem[{Sari et~al.(2017)Sari, Vlachos, and Stevenson}]{sari-etal-2017-continuous}
Yunita Sari, Andreas Vlachos, and Mark Stevenson. 2017.
\newblock \href {https://aclanthology.org/E17-2043/} {Continuous n-gram representations for authorship attribution}.
\newblock In \emph{Proceedings of the 15th Conference of the {E}uropean Chapter of the Association for Computational Linguistics: Volume 2, Short Papers}, pages 267--273, Valencia, Spain. Association for Computational Linguistics.

\bibitem[{Shi et~al.(2023)Shi, Wang, Yin, Chen, Chang, and Hsieh}]{shi2023redteaminglanguagemodel}
Zhouxing Shi, Yihan Wang, Fan Yin, Xiangning Chen, Kai-Wei Chang, and Cho-Jui Hsieh. 2023.
\newblock \href {https://arxiv.org/abs/2305.19713} {Red teaming language model detectors with language models}.
\newblock \emph{Preprint}, arXiv:2305.19713.

\bibitem[{Stamatatos(2009)}]{stamatatos2009survey}
Efstathios Stamatatos. 2009.
\newblock A survey of modern authorship attribution methods.
\newblock \emph{J. Am. Soc. Inf. Sci. Technol.}, 60(3):538–556.

\bibitem[{Williams et~al.(2024)Williams, Burke-Moore, Chan, Enock, Nanni, Sippy, Chung, Gabasova, Hackenburg, and Bright}]{williams2024largelanguagemodelsconsistently}
Angus~R. Williams, Liam Burke-Moore, Ryan Sze-Yin Chan, Florence~E. Enock, Federico Nanni, Tvesha Sippy, Yi-Ling Chung, Evelina Gabasova, Kobi Hackenburg, and Jonathan Bright. 2024.
\newblock \href {https://arxiv.org/abs/2408.06731} {Large language models can consistently generate high-quality content for election disinformation operations}.
\newblock \emph{Preprint}, arXiv:2408.06731.

\bibitem[{Xu et~al.(2024)Xu, Li, Xu, and Zhang}]{xu2024llmsknowrespectcopyright}
Jialiang Xu, Shenglan Li, Zhaozhuo Xu, and Denghui Zhang. 2024.
\newblock \href {https://arxiv.org/abs/2411.01136} {Do llms know to respect copyright notice?}
\newblock \emph{Preprint}, arXiv:2411.01136.

\bibitem[{Yoo et~al.(2023)Yoo, Ahn, Jang, and Kwak}]{yoo-etal-2023-robust}
KiYoon Yoo, Wonhyuk Ahn, Jiho Jang, and Nojun Kwak. 2023.
\newblock \href {https://arxiv.org/abs/2305.01904} {Robust multi-bit natural language watermarking through invariant features}.
\newblock \emph{Preprint}, arXiv:2305.01904.

\bibitem[{Zhang et~al.(2020)Zhang, Zhao, Saleh, and Liu}]{zhang2020pegasuspretrainingextractedgapsentences}
Jingqing Zhang, Yao Zhao, Mohammad Saleh, and Peter~J. Liu. 2020.
\newblock \href {https://arxiv.org/abs/1912.08777} {Pegasus: Pre-training with extracted gap-sentences for abstractive summarization}.
\newblock \emph{Preprint}, arXiv:1912.08777.

\bibitem[{Zhao et~al.(2023)Zhao, Ananth, Li, and Wang}]{zhao2023provablerobustwatermarkingaigenerated}
Xuandong Zhao, Prabhanjan Ananth, Lei Li, and Yu-Xiang Wang. 2023.
\newblock \href {https://arxiv.org/abs/2306.17439} {Provable robust watermarking for ai-generated text}.
\newblock \emph{Preprint}, arXiv:2306.17439.

\end{thebibliography}

\appendix
\section{Appendix}
\label{sec:appendix}

\subsection{Prompt for text generation with watermark}
\label{app:gen_prompt}

\begin{lstlisting}[style=prompt, numbers=none]
AMR (Abstract Meaning Representation) is a graph-based representation of a sentence's meaning. Each node is a concept and edges represent semantic roles or relationships. Below are some examples of template AMRs and corresponding sentences:
{example_text}

In the provided AMR, there are placeholders:
- "NE" for named entities (e.g., "Alice", "France", "Google").
- "N" for generic nouns (e.g., "a device", "an object").
- "X" for unspecified concepts (e.g., "something", "an idea").

Instructions:
- Do not write "NE", "N", or "X" literally. Instead, replace them with appropriate English words to form a natural, meaningful sentence.
- Ensure the generated sentence aligns with both the AMR structure and the given context.
- Do not produce multiple sentences or lists.
- Produce exactly one coherent sentence.

AMR:
{chosen_template}
Context: {context}
Please output only that one sentence.
\end{lstlisting}

\subsection{Prompt for zero-shot Claude paraphrase}
\label{app:para_prompt}
\begin{lstlisting}[style=prompt, numbers=none]
Previous context: {' '.join(context)}
Current sentence to paraphrase: {sent}
Rewrite the sentence above while preserving its meaning.
Do not provide any explanation or extra commentary.
Return only the new sentence.
\end{lstlisting}

\subsection{Prompt for LLM-as-a-Judge Text Quality Evaluation}
\label{app:quality_prompt}
\begin{lstlisting}[style=prompt, numbers=none]
You are an expert writing quality evaluator.

You will assess a GENERATED PARAGRAPH using the following criteria. For each, assign a score from 1 to 5 (decimals allowed), using the descriptions below.

1. **Coherence**: Measures how logically and clearly the ideas are organized and connected.
   - 1: Incoherent; sentences are unrelated or confusing.
   - 2: Poor transitions or unclear relationships between ideas.
   - 3: Basic logical flow, but some awkward connections.
   - 4: Mostly logical and clear, with minor lapses.
   - 5: Highly logical and seamless flow of ideas.

2. **Fluency**: Assesses the grammatical correctness and naturalness of the language.
   - 1: Grammatically broken or unreadable.
   - 2: Understandable but awkward or error-prone.
   - 3: Generally readable, some minor grammatical errors or odd phrasing.
   - 4: Well-written with only occasional issues.
   - 5: Grammatically correct and naturally flowing throughout.

3. **Diversity**: Use of varied vocabulary and sentence structure, avoiding repetition.
   - 1: Extremely repetitive or formulaic.
   - 2: Some repetition with occasional variation.
   - 3: Moderate variety; not monotonous.
   - 4: Good diversity in language and structure.
   - 5: Highly expressive and varied without redundancy.

**Scoring Instructions**:
- Return a score for each of the three dimensions above.
- You may use decimal values (e.g., 2.5, 4.7).

**Output Format**:
Respond with a **valid JSON object only** in this exact format:

{{
  "coherence_score":  float,
  "fluency_score":    float,
  "diversity_score":  float
}}
\end{lstlisting}

\end{document}